# AC-Driven Series Elastic Electrohydraulic Actuator for Stable and Smooth Displacement Output

Quan Xiong, *Student Member*, *IEEE*, Xuanyi Zhou, Dannuo Li and Raye Chen-Hua Yeow, *Member*, *IEEE*

*Abstract*—soft electrohydraulic actuators known as HASEL actuators have attracted widespread research interest due to their outstanding dynamic performance and high output power. However, the displacement of electrohydraulic actuators usually declines with time under constant DC voltage, which hampers its prospective application. A mathematical model is firstly established to not only explain the decrease in displacement under DC voltage but also predict the relatively stable displacement with oscillation under AC square wave voltage. The mathematical model is validated since the actual displacement confirms the trend observed by our model. To smooth the displacement oscillation introduced by AC voltage, a serial elastic component is incorporated to form a SE-HASEL actuator. A feedback control with a proportion-integration algorithm enables the SE-HASEL actuator to eliminate the obstinate displacement hysteresis. Our results revealed that, through our methodology, the SE-HASEL actuator can give stable and smooth displacement and is capable of absorbing external impact disturbance simultaneously. A rotary joint based on the SE-HASEL actuator is developed to reflect its possibility to generate a common rotary motion for wide robotic applications. More importantly, this paper also proposes a highly accurate needle biopsy robot that can be utilized in MRI-guide surgical procedures. Overall, we have achieved AC-driven series elastic electrohydraulic actuators that can exhibit stable and smooth displacement output.

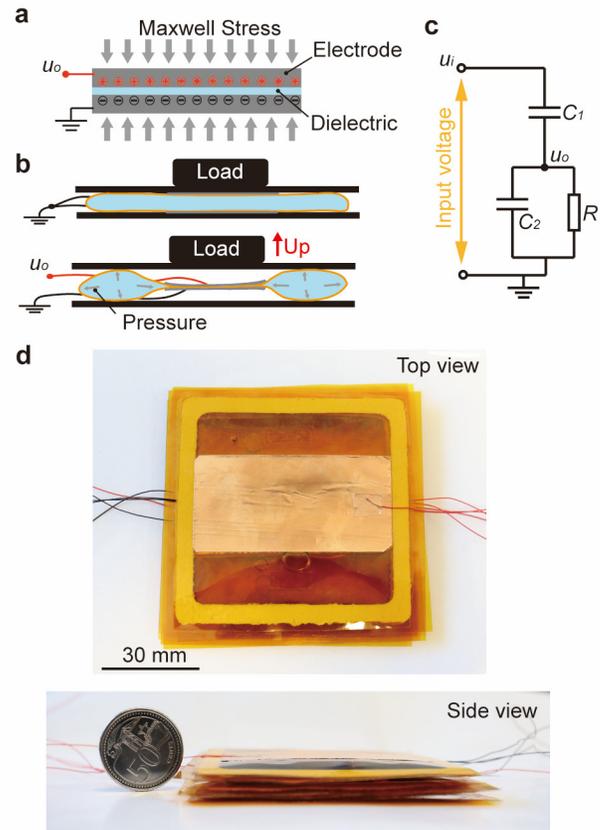

**Fig. 1.** The soft electrohydraulic actuator. (a) The principle of electrostatic adhesion. (b) The principle of electrohydraulic actuators. (c) The leakage model of EA. (d) The actual electrohydraulic actuators made of polyimide films and dielectric oil.

## I. INTRODUCTION

SOFT actuators demonstrated tremendous potential in extensive robotic applications, such as grasping delicate objects with low risk of damage [1]–[4], interacting safely with humans for assistance and rehabilitation [5]–[7], and maneuvering in highly confined environments [8]–[10]. Compared with conventional rigid actuators like geared motors, soft actuators possess a compliant and deformable feature that brings about high safety and adaptability to soft robotic systems [11]–[13]. Currently, soft pneumatic actuators as the mainstream soft actuators have been developed for various soft robots like prosthetic robots, surgical robots, exoskeletons and locomotion robots [2], [14]–[17]. However, these require indispensable gas pressure or vacuum electromagnetic motor-based pumps often limiting their performance and portability [18], [19].

The emerging flexible electrostatic adhesion (EA) technology has been introduced into soft robotics in these years [20]–[23]. For example, soft grippers made of membranous EA can promptly stick to objects with low power consumption and low contacting stress to objects and unstick from objects by just powering off [24]–[26]. The flexible EA technologies were also integrated into haptic devices that can generate desired feedback force to human fingers by regulating applied voltages [27]–[29].

Recently, the soft actuators based on flexible EA attracted

This work was supported by National Robotics Programme–Robotics Enabling Capabilities and Technologies (W2025d0243) and A*STAR Industry Alignment Fund – Pre-Positioning (A20H8A0241). *(Corresponding author: Raye Chen-Hua Yeow).*

Quan Xiong, Xuanyi Zhou, Dannuo Li and Raye Chen-Hua Yeow are with the Evolution Innovation Lab, Department of Biomedical Engineering, National University of Singapore, Singapore, 117583 Singapore (e-mail: e0788090@u.nus.edu; zhou447837285@gmail.com; ldannuo@gmail.com; rayeow@nus.edu.sg).

Raye Chen-Hua Yeow is also with Computer Science & Artificial Intelligence Laboratory, Massachusetts Institute of Technology, 02139, United States.

much attention from researchers since they paved a prospective path to portable and soft actuators without tethered connection to sources of pressure or vacuum fluid [18], [30]. Dielectric elastomer actuators (DEA) make use of EA force to squeeze the dielectric elastomer and generate mechanical deformation. Although DEAs have widespread soft robotic applications like micro flapping robots and crawling robots [8], [31]–[33], the main disadvantage lies within their low tolerance to electric breakdown which can result in severe damage. Moreover, their pre-stretched dielectric elastomer films also make their fabrication process complicated. To surmount those above challenges, electrohydraulic actuators known as hydraulically amplified self-healing electrostatic (HASEL) actuators were presented [18], [34], [35]. A dielectric liquid bladder is substituted for the dielectric elastomer in HASEL actuators, and the EA between two electrodes deforms the bladder and actuates the load. They displayed excellent performance including high power/weight ratio, fast response, and low-cost manufacturing process.

More importantly, conventional pneumatic actuators with pumps and electromagnetic motors are inadequate in magnetic-sensitive scenarios, such as surgeries or medical examinations within magnetic resonance imaging (MRI) devices [36]–[38]. The strong electromagnetic fields will affect the imaging quality, or even damage the MRI device. Nevertheless, EA-based actuators without permanent magnets can work as normal in the MRI environment [39], since they only generate negligible magnetic interference due to their extreme-low current (μA level).

Constant DC voltages have been widely applied to electrohydraulic actuators [35], [40], [41]. It is a common phenomenon that DC operating voltage causes gradual EA force drops with time [42], [43]. This phenomenon more or less occurs in many EA devices with a variety of dielectric materials (like Biaxially-oriented Polypropylene (BOPP), polyethylene terephthalate (PET), polyimide (PI), Mylar, Polyvinylidene Fluoride (PVDF), Luxprint, et al.), such as EA clutches [27], [44], [45] and electrohydraulic actuators [40], [46]–[48]. Despite the possibility of selecting appropriate dielectric materials with less EA force loss since the EA decreasing rates under DC voltage are different in materials, it may sacrifice other significant attributes, like mechanical behavior and temperature variations.

In this work, we proposed to use low-frequency AC square wave voltage to drive the electrohydraulic actuators. Using AC operating voltage can eliminate the EA force decrease and produce stable displacement output. However, it also led to periodical oscillation which is undesirable. We then connected an elastic component in series with the electrohydraulic actuators to isolate this oscillation, i.e., series elastic HASEL (SE-HASEL) actuators. Even though the resultant displacement of SE-HASEL actuators is stable and smooth, obvious hysteresis due to large magnitudes of applied AC voltage was observed in our following experiments, which dims the prospects for model-based open-loop control. Thus, we adopted to implement feedback control on our actuators.

We characterized the output displacement decrease under DC voltage using different electrohydraulic actuators and the amplitudes of displacement oscillation under different frequencies of AC voltage and loads. The oscillation isolation effect with our presented SE-HASEL actuators was also verified by experiments. Furthermore, experimental results indicated that our actuators with feedback control can overcome hysteresis and achieve accurate, stable and smooth displacement outputs. We also developed a rotary robotic joint and a precise needle biopsy robot driven by SE-HASEL actuators, which could potentially be deployed in MRI-guided surgical procedures.

The main contributions of this article are as follows. First, we built a mathematical model to explain the EA force decline and predict the oscillation resulting from AC square wave voltage based on a leakage model. Second, we proposed the SE-HASEL actuators with AC operating voltage capable of outputting stable and smooth displacement. Third, we reported a novel fabrication process for electrohydraulic actuators made of polyimide (PI) membranes which have high mechanical strength and good temperature tolerance. Fourth, a robust feedback controller is implemented to improve the control accuracy. Last, we are the first to deploy electrohydraulic actuators for medical biopsy.

## II. MODELING

A typical EA device is composed of two electrodes and a sandwiched layer of dielectric material. Applying high voltage to the electrodes, positive charges are induced to the positive electrode while negative charges accumulate to the negative counterpart (see Fig. 1a). The Maxwell stress generated by Coulomb force of charges with different polarity (Electrostatic adhesion) vertically squeezes the two electrodes. The EA can be calculated based on the parallel capacitor model as follows,

$$F = \frac{1}{2}\frac{\varepsilon_r \varepsilon_0 A u_o(t)^2}{d^2}, \qquad (1)$$

where $u_o(t)$ is the actual applied voltage, $\varepsilon_r$ is the relative permittivity of the insulation layer, $\varepsilon_0$ is the vacuum permittivity, $A$ is the overlap area of the two electrodes, and $d$ is the thickness of the dielectric layer.

The electrohydraulic actuator replaces the dielectric layer with a deformable dielectric bladder made of a dielectric membranous shell and inner dielectric liquid. The flexible electrodes compress the bladder by EA and cause its deformation which lifts mechanical loads (see Fig. 1 b and d).

### A. Electrostatic Adhesion under DC Voltage

The decline of EA under constant DC applied voltage was indicated in many pieces of research. An electric leakage model [49] was built to explain the decline of EA in haptic applications (see Fig. 1c). We also use this model for our electrohydraulic actuators. A leakage resistance, $R$ is connected in parallel to the two electrodes (capacitor $C_2$) of the electrohydraulic actuator, and another capacitor $C_1$ in the cable of input voltage $u_i$ (see Fig. 1c). The actual applied voltage to the electrohydraulic actuator is the output voltage $u_o$. To build the mathematical model of the leakage model, we first derive a differential equation by Kirchhoff's law,

$$C_1 \frac{d(u_i - u_o)}{dt} = C_2 \frac{du_o}{dt} + \frac{u_o}{R}. \quad (2)$$

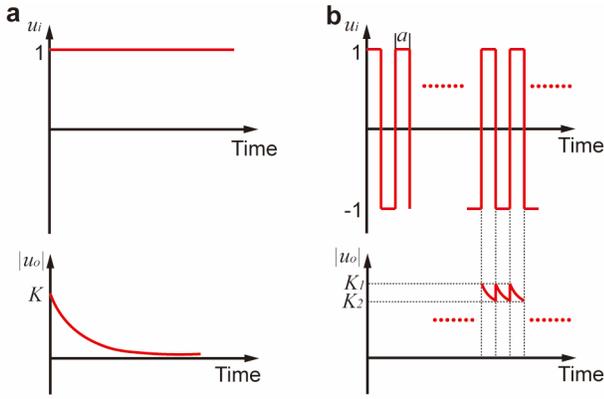

**Fig. 2.** The input voltage and output voltage based on leakage model. (a) The response under constant DC voltage. (b) The response under AC square wave voltage.

Then we get the transfer function by Laplace transformation,

$$T(s) = \frac{U_o(s)}{U_i(s)} = \frac{RC_1 s}{R(C_1 + C_2)s + 1}. \quad (3)$$

Considering the constant DC input voltage as the unit step signal $1(t)$, we can calculate the input voltage $U_i$ and the output voltage $U_o$ in the Laplace domain,

$$U_i = \frac{1}{s}, \quad (4)$$

$$U_o = T(s)U_i = \frac{RC_1}{R(C_1+C_2)s+1}. \quad (5)$$

The actual applied voltage in the time domain is also calculated by inverse Laplace transformation,

$$u_o(t) = Ke^{Pt}1(t). \quad (6)$$

$$K = \frac{C_1}{C_1 + C_2}, \quad (7)$$

$$P = \frac{-1}{R(C_1 + C_2)}, \quad (8)$$

Where $K$ and $P$ are constants. Hence, the actual applied voltage reduces to 0 V with time under a constant DC voltage as shown in Fig. 2a.

*B. Electrostatic Adhesion under AC Square Wave Voltage*

We then establish the mathematical model of the actual output voltage under AC square wave input voltage (see Fig. 2b). We assume the magnitude of the AC voltage is 1 V, and the frequency is $f$. The input voltage described in the time domain is

$$u_i(t) = 1(t) + 2\sum_{k=1}^{n}(-1)^k 1(t-ka), \quad (9)$$

$$t = \Delta t + na \quad \Delta t \in [0, a) \quad n \in N, \quad (10)$$

$$a = \frac{1}{2f}, \quad (11)$$

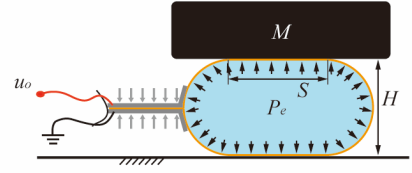

**Fig. 3.** The part section view of the deformed bladder of electrohydraulic actuator.

where $t$ is the actual time, $a$ is the time interval when the input voltage remains constant, and $N$ represents the natural numbers. The input voltage $U_i$ and output voltage $U_o$ can be calculated in Laplace domain,

$$U_i = \frac{1}{s} + \frac{2}{s}\sum_{k=1}^{n}(-1)^k e^{-kas}, \quad (12)$$

$$U_o = T(s)U_i = \frac{RC_1}{R(C_1+C_2)s+1}(1+2\sum_{k=1}^{n}(-1)^k e^{-kas}). \quad (13)$$

We can inversely transfer them to time domain functions,

$$u_o(t) = Ke^{Pt} + \sum_{k=1}^{n} 2K(-1)^k e^{P(t-ka)}$$
$$= Ke^{Pt} + \frac{1}{1+e^{Pa}}(-2Ke^{Pt} + 2K(-1)^n e^{P(t-na)}) \quad (14)$$

If time is infinite ($n$ is infinite), according to (10) and (14), the output voltage is

$$\lim_{n \to +\infty} u_o(t) = (-1)^n 2Ke^{P\Delta t}\frac{1}{1+e^{Pa}}. \quad (15)$$

Noted that the absolute value of output voltage oscillates periodically with time (at twice the frequency of AC input voltage), but remains roughly stable with time, according to (15), as shown in Fig.2b, where $K_1$ and $K_2$ are the upper and lower limitations respectively,

$$K_1 = \frac{2K}{1+e^{Pa}}, \quad (16)$$

$$K_2 = \frac{2Ke^{Pa}}{1+e^{Pa}}. \quad (17)$$

This voltage oscillation occurs when the polarity of the input voltage is reversed, and meanwhile results in the sudden change of EA force based on (1).

*C. Displacement of Electrohydraulic Actuators*

A mathematical model is also required to describe the relationship between the EA and the displacement output of electrohydraulic actuators. Applying a high voltage to the electrodes, part electrodes zip by EA, and squeeze the dielectric liquid to the other side. Meanwhile, the movement of the liquid lifts the load $M$. The section view of the activated electrohydraulic actuator is demonstrated in Fig. 3. Generally, the stiffness of thin dielectric films and flexible electrodes (micro-meter scale) are neglected and the liquid is incompressible. The volume of liquid is constant and the dielectric membranes on both sides of the bladder are circular so that we can derive,

$$(HS + \pi(H/2)^2)L_0 = V_0. \quad (18)$$

Where $H$ is the displacement output of the electrohydraulic actuator, $S$ is the contacting length, $L_0$ and $V_0$ are constants, that are the width of the electrohydraulic actuator and the volume of liquid respectively. Given the equilibrium conditions, the fluidic pressure is related to the load,

$$L_0 P_e S = Mg, \quad (19)$$

where $P_e$ is the pressure of the dielectric liquid and $Mg$ is the load. We can calculate the displacement,

$$H = \frac{2}{\pi}\sqrt{(\frac{Mg}{L_0 P_e})^2 + \frac{\pi V_0}{L_0} - \frac{2Mg}{\pi L_0 P_e}}. \quad (20)$$

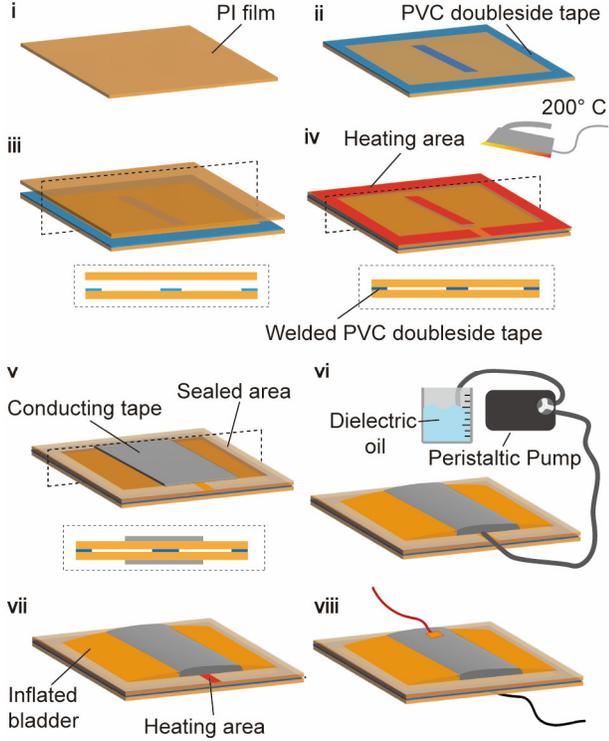

**Fig. 4.** The fabrication of electrohydraulic actuators made of polyimide films.

Considering that the electrodes are not completely zipped with load, we can acquire another equilibrium relationship between the liquid pressure and EA stress, as follows,

$$P_e = \frac{1}{2}\frac{\varepsilon_{film}\varepsilon_0 u_o(t)^2}{d_{film}^2}, \quad (21)$$

where $\varepsilon_{film}$ is the relative permittivity of the dielectric bladder shell and $d_{film}$ is the total thickness of the double-fold dielectric films (the zipped bladder). Combined with the oscillation of the output voltage due to AC input voltage and this displacement model, we can predict the output displacement oscillation of the electrohydraulic actuator under AC square wave voltage.

### III. FABRICATION AND SYSTEM ASSEMBLY

#### A. Fabrication of electrohydraulic actuators

Polyimide (PI) materials with a high relative permittivity (3.4 which is approximately 50% higher than BOPP materials), low cost, good mechanical strength and

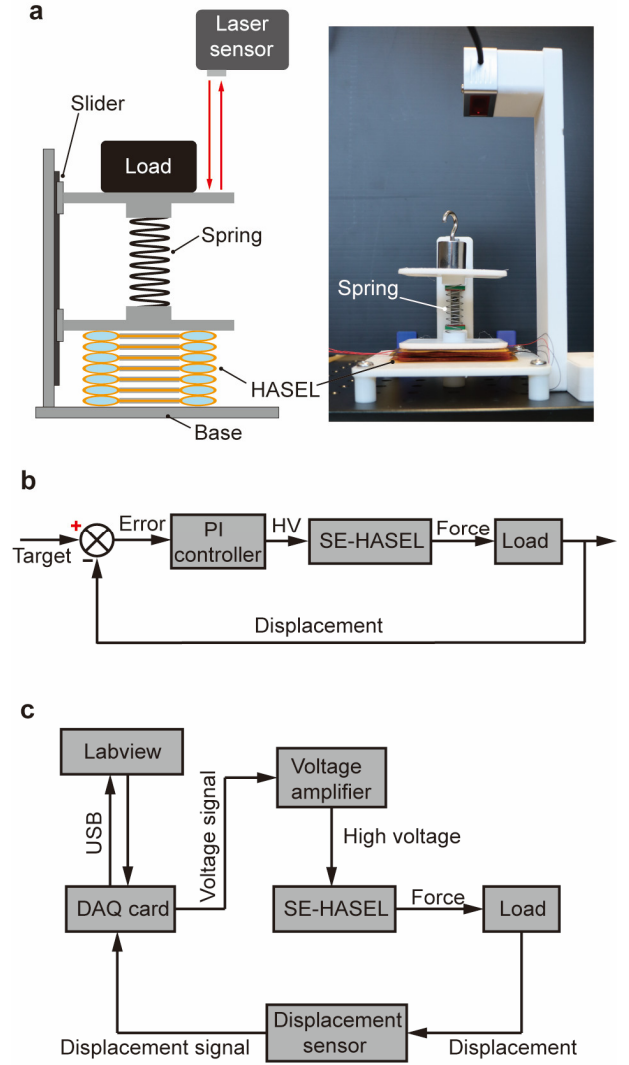

**Fig. 5.** The system composition. (a) The mechanical design and prototype of SE-HASEL actuator. (b) The feedback control diagram. (c) The electronic block diagram.

temperature tolerance, have been widely used in electric industries, like FPC (flexible printed circuits) based on PI membranes. However, it is hardly used in electrohydraulic actuators, mainly because of its complicated and expensive adhesive bonding technique for enclosed dielectric bladders. Here, we proposed a low-cost adhesive bonding method by using PVC double-adhesive tapes to fabricate the electrohydraulic actuators. First, we cut the PVC tape sandwiched by two release papers to a box shape (6×6 cm of the inner square and 7×7 cm of the outer square) by a cutting machine (Silhouette Cameo, Silhouette). Second, we stick the cut PVC tape to a PI film (50 μm thick) and stick a narrow tape (3 mm wide) in the middle to divide it into two chambers. Third, another PI film covers the other side but leaves a narrow channel by inserting a scrap of release paper for liquid import. Fourth, we use an iron (200 °C) to heat the PVC tape area except for the inlet channel (see Fig. 4 i to iv) to bond two

PI films. Fifth, we put two copper tapes in the middle of the PI bladder as electrodes. Sixth, 3.2 mL dielectric oil is imported into the bladder by a peristaltic pump (YX-LP01-3, Yanxiao). Last, we seal the import channel with the heating iron in the fourth step and then taped wires to electrodes (see Fig. 4 v to viii).

*B. SE-HASEL System*

To smooth the output displacement of electrohydraulic actuators, a spring is connected with a stack of electrohydraulic actuators in series (SE-HASEL actuator). We used two vertical sliders to constrain the direction of displacement (see Fig. 5a). A plate mounted on the lower slider is in contact with electrohydraulic actuators, and the other plate mounted on the upper slider holds the load. The damping of sliders, the spring and the mass compose a typical second-order system, which acts as a passive low-pass filter to restrain the oscillation pulse from electrohydraulic actuators (oscillation isolation).

*C. Feedback Control*

A Proportion-Integration controller and the feedback control strategy (see Fig. 5b) are applied to overcome the hysteresis of output displacement. The Proportion-Integration algorithm real-time calculates the magnitude of the AC square wave voltage,

$$|u_i| = K_p e + K_i \int e \, dt . \quad (22)$$

where $K_p$ is the proportion parameter, $K_i$ is the integration parameter and $e$ is the error value between the target and actual output displacement of the SE-HASEL actuator (the upper plate in Fig. 5a). The frequency of the AC applied voltage is pre-set. $K_p$ and $K_i$ are manually pre-adjusted.

*D. Electric System*

The electronic block diagram is indicated in Fig. 5c. A laser sensor (BL-100NMZ, BOJKE) is mounted over the upper plate (see Fig. 5a) to measure the actual output displacement of the SE-HASEL actuator. A data acquisition device (USB6211, NI-DAQ) acquires this displacement signal from the laser sensor and transfers it to the LabVIEW platform (LabVIEW 2018) on a local desktop computer. Our Proportion-Integration control algorithm embedded in the LabVIEW platform calculates the needed magnitude of AC square wave voltage in real-time. Then, the LabVIEW platform modulates this magnitude value to an AC square wave and returns the modulated AC voltage signal to the DAQ which outputs the AC voltage to a voltage amplifier (Trek 10/40, Advanced Energy). The AC voltage is amplified by 1000 times and then supplied to the SE-HASEL actuator.

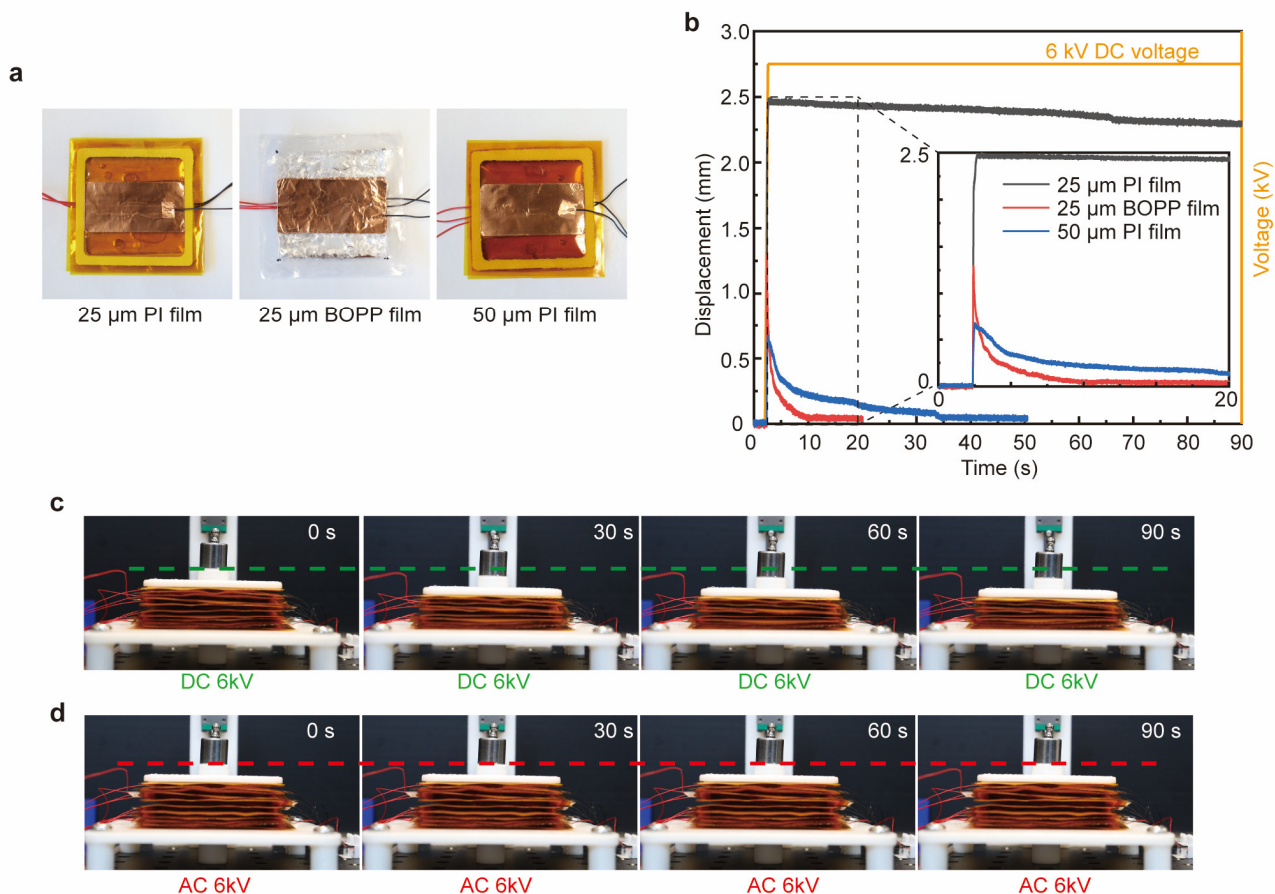

**Fig. 6.** The displacement decrease under DC voltage. (a) Three different electrohydraulic actuators. (b) The actual output displacements of three electrohydraulic actuators made of three different materials under a 6 kV constant DC voltage. The snapshots of electrohydraulic actuators: (c) Under constant DC voltage. (d) Under AC square wave voltage.

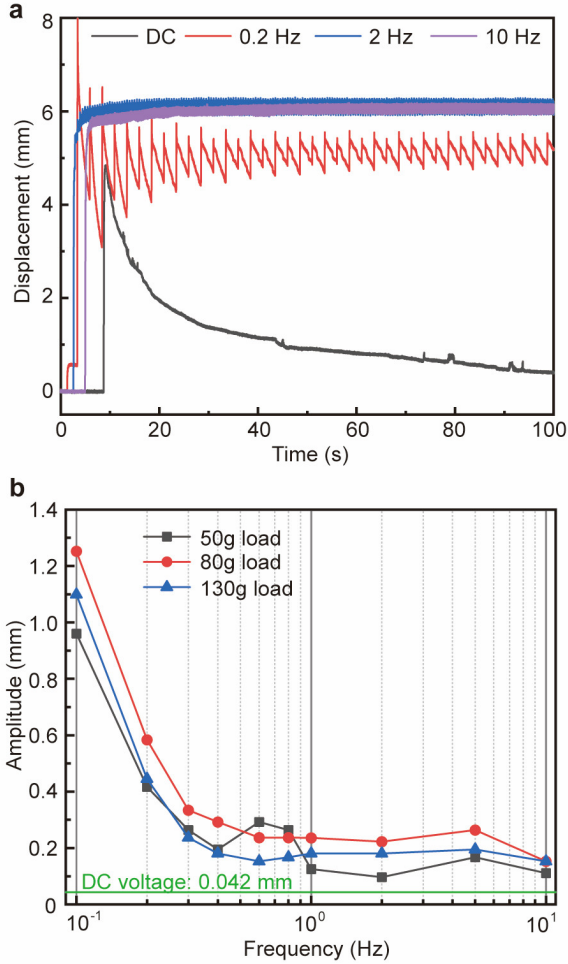

**Fig. 7.** Displacement under AC Voltage. (a) The actual output displacements under the AC square wave voltages with different frequencies and constant DC voltage. (b) The oscillation amplitude of the output displacement under the AC square wave voltage with different frequencies and different loads.

IV. EXPERIMENTS AND DISCUSSION

*A. Displacement Decrease under DC Voltage*

We experimentally explored the drop of output displacement of electrohydraulic actuators (stacking 11 actuators) under constant DC voltage. First, we built a vertical linear slider setup to visually observe the output displacement. A 20 g weight was fixed on the plate which was mounted on the sliding block (30 g). While supplying a constant 6 kV DC voltage to the actuators, the output displacement leaped up rapidly at the beginning and then declined gradually (see Fig. 6c and video S1).

To quantitively characterize the effect of using different dielectric materials on displacement, we manufactured three electrohydraulic actuators (3 stacked electrohydraulic actuators) with 25 μm thick PI films, 25 μm thick pre-coated BOPP films [50] and 50 μm thick PI films with the same size and the same volume of liquid as described in Section III.A (see Fig. 6a). We applied a constant 6 kV voltage to the three actuators and used a laser sensor to measure the actual output displacement. The rate at which displacement decreased varied greatly with the type of dielectric materials used as illustrated in Fig. 6b. The actuators made of 25 μm PI films remained relatively stable with time as their displacement reduced by only 6.5% in 80 seconds from applying the voltage. Nevertheless, the displacement of pre-coated BOPP films dropped dramatically by 96.3% in 10 seconds. The displacement of 50 μm PI films showed an intermediate drop (71% in 10 seconds). Thus, we used the 50 μm thick PI film material in our following experiments. Overall, the displacement drop can be explained by our mathematical model (6).

*B. Oscillation of Displacement Output under AC Voltage*

To keep the EA force stable, we adopted the AC square wave applied voltage to our electrohydraulic actuators. In contrast with the displacement output under DC voltage, it always maintained an almost constant value with visible oscillation under an AC square wave voltage with 6 kV magnitude (see Fig. 6d and video S1). Then we altered the frequency of the applied voltage (DC, 0.2 Hz, 2 Hz and 10 Hz), and measured the actual displacements by the displacement sensor. The experimental results are shown in Fig. 7a. With the AC square wave applied voltage, the displacement output was generally stable over time and oscillating around the stable value at twice the frequency of AC applied voltage (clearly observed in the red curve in Fig. 7a), which was exactly revealed by our mathematical model as shown in Fig. 2b, (20) and (21). With a frequency from 0.2 Hz to 2 Hz, the stable value under AC applied voltage increased obviously, but it kept nearly unchanged when the frequency was over 2 Hz.

We further investigated the displacement oscillation by testing the displacement amplitude with different frequencies from 0.1 Hz to 10 Hz and three different loads (50 g, 80 g and 130 g). We collected the displacement data and manually calculated the amplitude using the peak-peak value at the stable stage (60 s after applying voltage). The results showed that the amplitudes declined with the frequency of the applied voltage (see Fig. 7b), which was consistent with the trend predicted by our mathematical model as shown in (16) and (17). By increasing the time interval *a* (reducing frequency) of AC voltage, the difference between the upper limit $K_1$ and the lower limit $K_2$ rises along with the amplitude. In addition, the load also influenced the amplitude. The amplitude of the 80 g was observed to be higher than the 50 g and 130 g loads.

Noted that, the decline of amplitude with the frequency levels off, as the frequency passes 1 Hz. Hence, we can apply an AC voltage with a relatively lower frequency (2 Hz) to our SE-HASEL actuators. There are four main advantages as follows. First, a 2 Hz AC voltage is sufficient to maintain a stable displacement output with acceptable oscillation. Second, the power consumption augments with the frequency. Third, a higher frequency of voltage is more hazardous to the human body [51]. Last, the higher frequency increases the hardware requirements and the cost of electronic components for portable

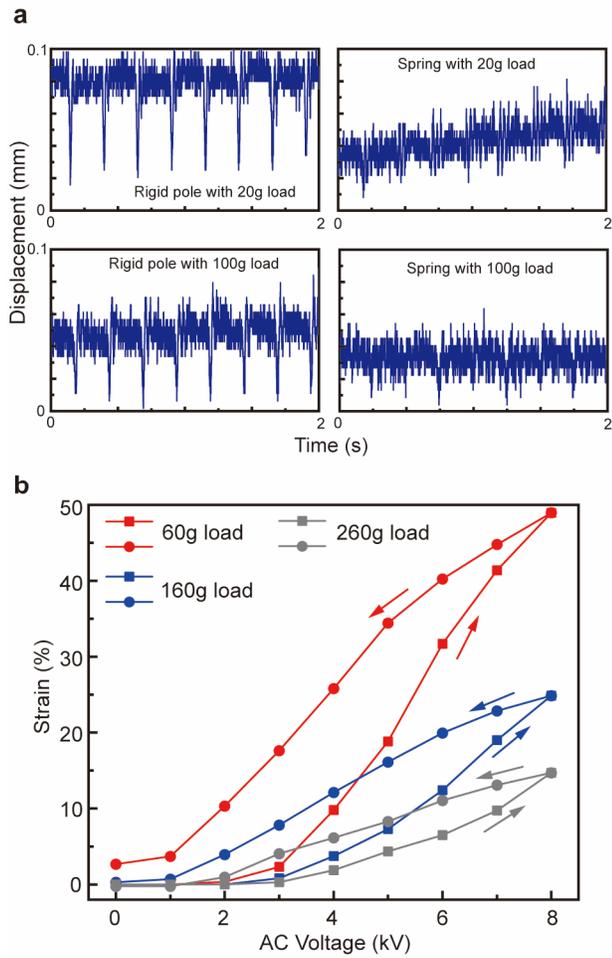

**Fig. 8.** Displacement output of the SE-HASEL actuator. (a) The displacement output of the electrohydraulic actuators and our proposed SE-HASEL actuator. (b) The displacement hysteresis with the magnitude of AC applied voltage.

device development (e.g., the faster switching response of high voltage relays and optocouplers).

*C. Displacement Output of the SE-HASEL Actuator*

Although the oscillation can be reduced to a relatively low level by increasing frequency, the periodical vibration is also undesirable, especially in some delicate robotic operations. To achieve smoother displacement output, we used a series spring (14 N/m stiffness) to isolate the oscillation from the electrohydraulic actuators (SE-HASEL actuator). We applied a 2 Hz AC square voltage with 6 kV magnitude to the SE-HASEL actuator with a 20 g weight on the upper plate and recorded the actual displacement. Then, we substituted the spring with a rigid pole and re-conducted the testing. Furthermore, with a 100 g weight, we also tested the actual displacement with a spring and a rigid pole. The results indicated that SE-HASEL actuators can output a smooth displacement without obvious impulse (see Fig. 8a).

We also experimentally characterized hysteresis behavior by varying the magnitude of AC applied voltage from 0 to 8 kV. SE-HASEL actuators cannot eliminate the output displacement hysteresis completely. We tested the output displacement with three loads (60 g, 160 g and 260 g), and calculated the strain (dividing the output displacement by the total height of the stack of electrohydraulic actuators) as shown in Fig. 8b. With a 60 g load, the maximum hysterical strain (MHS) is about 15% at 3 kV applied AC voltage. The maximum output strain (MOS) is about 50 %. The ratio of MHS to MOS is 30%. The hysteresis subsides and the MOS also declines with heavier load. However, the ratio of MHS to MOS was roughly maintained at 30% with different loads.

*D. Displacement Output with Feedback Control*

We applied the feedback control to our SE-HASEL actuator (11 stacked electrohydraulic actuators) with a Proportion-Integration controller. The control cycle was set to 1 ms in the LabVIEW platform, and the controller parameters $K_p$ and $K_i$ were manually adjusted to 0.8 and 0.005. We put a 50 g weight on the SE-HASEL actuator, and then implemented the feedback control to follow the target waves (0.05 Hz square waves and sine waves). The displacement and applied voltage were recorded in Fig. 9a. For the square waves with a constant offset, the displacement of our SE-HASEL actuator can stably and smoothly follow the target with less than 10 % overshoot, and the root-mean-square error (RMSE) is 0.077 mm. At the falling edges, the actual displacement usually lags behind the target, mainly because the electrohydraulic actuator can only generate the unidirectional force (upward). The sinking of the load was only dominated by its gravities.

Our SE-HASEL actuator with PI feedback control can also follow the sine waves with slight fluctuation and stick-slip phenomenon which is probably due to the nonlinear friction during movement. However, the following performance at low target values is poor resulting in the total RMSE (0.211 mm) rising to 3 times that of square waves. The magnitude of applied AC voltage was effectively modulated by the Proportion-Integration controller to correct the error between the actual value and the target value (see Fig. 9b).

Then target waves with higher frequencies (0.1 Hz, 0.2 Hz and 0.5 Hz) were also tested respectively as shown in Fig. 9 c and d. Overall, the actual displacement can follow the under 0.2 Hz target waves without attenuation and phase lag. For the square waves, the maximum actual value overshooted the target value by 13 %, and the actual displacement at the falling stage also lagged behind the target. The occasionally large overshoot at the falling edge is mainly because of the sudden release of the gravitational potential energy of the lifted load. The output displacement cannot reach the up and down plateaus of the 0.5 Hz square wave. For the sine waves, less overshoot was observed, but the control precision at low target values also needed to improve. When the frequency was to 0.5 Hz, the actual displacement attenuated, but its phase anchored the target waves.

To demonstrate the vibration isolation of the SE-HASEL actuator, we obtained the actual displacement of the upper and lower plates (Plate A and Plate B, see Fig 10) under a feedback control process in real-time. We stuck two right paper plates to Plate A and Plate B, and two laser sensors were used to measure the displacement simultaneously. The displacement of Plate A which was isolated by the series spring (the output displacement of the SE-HASEL actuator) was smoother and controlled to a target value. However, the displacement of Plate B which was directly in contact with the electrohydraulic actuators oscillated periodically (sharp peaks in the curve, see Fig. 10 and video S2).

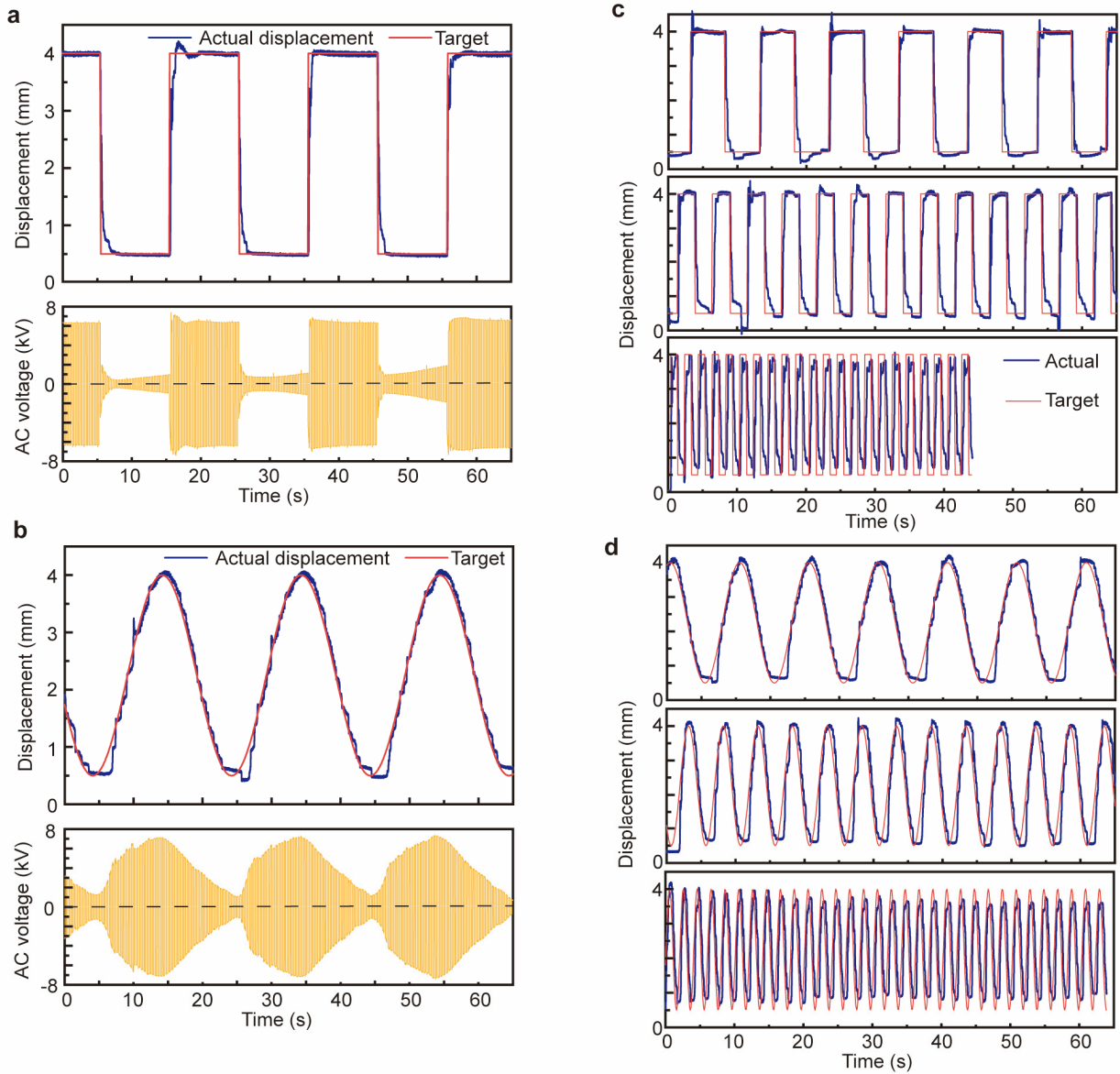

**Fig. 9.** The Displacement output with feedback control. The SE-HASEL actuator following targets waves and its actual applied voltage. (a) The square waves. (b) The sine waves; The SE-HASEL actuator following targets square and sine waves (0.1 Hz, 0.2 Hz and 0.5 Hz frequency). (c) The square waves. (d) The sine waves.

*E. Impact Absorption of the SE-HASEL Actuator*

The impact, disturbance, or variation of the load will dramatically influence the output displacement of electrohydraulic actuators, and even destabilize the system. Here, we conducted an impact disturbance test. First, the output displacement of the SE-HASEL actuator was controlled to 5 mm. Then, we applied an instantaneous 50 g weight on Plate A to simulate an impact disturbance. After the displacement stabilized, we removed the load. The actual displacement and the applied voltage are shown in Fig. 11 and video S3. After applying the load, the absolute displacement dropped rapidly to almost 1 mm and then recovered to 5 mm in 1 second when the magnitude of AC voltage was increased. Upon removing the load, the displacement soared to 11 mm

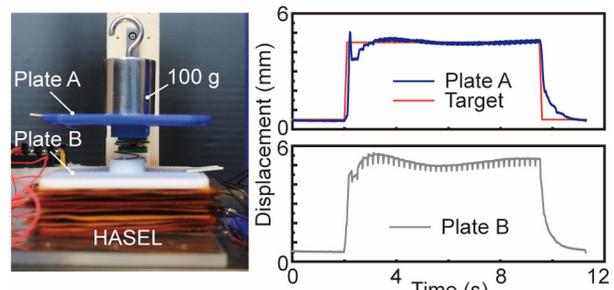

**Fig. 10.** The actual displacement of Plate A and Plate B in the SE-HASEL actuator.

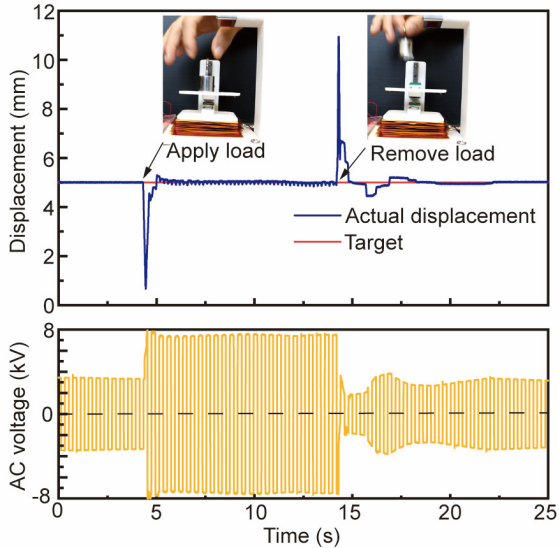

**Fig. 11.** The impact absorption of the SE-HASEL actuator with feedback control.

and quickly descended to 5 mm with slight fluctuation around this value.

The results revealed that the elastic component in the SE-HASEL actuator can not only isolate the vibration from the electrohydraulic actuators to the load but also buffer the external impact or disturbance from the load to the electrohydraulic actuators inversely. It allows the control system to unhurriedly and stably respond to the impact or disturbance.

### F. Applications
*1) Rotary Robotic Joint*

To demonstrate the application potential, we developed a rotary robotic joint based on the SE-HASEL actuator. We designed a link-slider mechanism to transmit the linear displacement to a rotation movement (see Fig. 12). We manually increased the target value from 0 mm to 8 mm by a 2 mm increment, and then inversed the process back to 0 mm. The snapshots in Fig. 12a and video S4 explicitly display the rotation movement (rotation angle range: 48.1 degrees) of the joint.

Though there are some sharp peaks in the output displacement curve (see Fig. 12b), probably due to the large moment of inertia of the joint and friction, the control system was generally stable and the actual displacement of the SE-HASEL actuator matched the arbitrary target with acceptable error (0.386 mm RMSE). After the 50s, the displacement cannot return to 0 rapidly, mainly because of the unidirectional actuation property of electrohydraulic actuators. At this moment, the applied voltage is already 0 V and only gravity cannot overcome the slight plastic deformation of electrohydraulic actuators.

*2) Needle Biopsy Robot*

The needle biopsy is an important medical method for cancer diagnosis and pathological analysis [52], which attracts many researchers to dive into needle biopsy robotics. An alternative technique (MRI) can provide cross-sectional images of the human body that can guide the biopsy robot and improve the penetration accuracy for needle biopsy [53]–[55]. Here, we showcased a needle biopsy robot developed by our SE-HASEL actuator, to clarify its particular compatibility with MRI scenarios where it is quite challenging for conventional electromagnetic motor-based needle biopsy robots.

We designed a needle fixture to replace Plate A in Fig. 10, and vertically mounted a syringe needle on this fixture (Fig. 13a). Over the needle head, we used a piece of chicken breast (about 2 mm thickness) and a piece of pork liver to simulate human muscle and visceral organ. A 3 mm thick plate was sandwiched by the chicken and liver leaving only a 0.9 mm air gap due to the deformation of the chicken and liver to simulate other human tissue (see Fig. 13 a and b). There is a 6 mm thick breach in this plate for needle penetration and observation. A 50 g weight was also fixed on the needle fixture to supply a recovering force for pushing the needle out (see the side view in Fig. 13b).

First, we adjusted the position of the needle head to the bottom of the muscle (chicken), and manually measured the actual distance from the needle head to our target position, which is similar to the guiding function of MRI. We then set the target displacement of the SE-HASEL actuator to achieve different biopsy samples: the liver tissue and the gap 'tissue'. For liver tissue sampling, we first set the target to 2 mm and then increased the target to 3.5 mm. The needle penetrated the muscle tissue, following the liver tissue (see video S5 and Fig. 13d). The actual displacement as shown in Fig. 13c can follow our targets with only 0.067 mm RMSE. For liver tissue sampling, we directly set the target to 2.5 mm. The needle penetrated the muscle tissue, but stayed within the gap 'tissue' (see video S5 and Fig. 13d). The actual displacement can also follow the target precisely (only 0.051 mm RMSE). The snapshots and the microscopic views of liver tissue in Fig. 13d clearly display the difference in the deepest needle position between these two samplings. Our SE-HASEL actuator with feedback control can output accurate, smooth and stable displacement, which can avoid additional damage to the human body and is crucial for medical robotic applications.

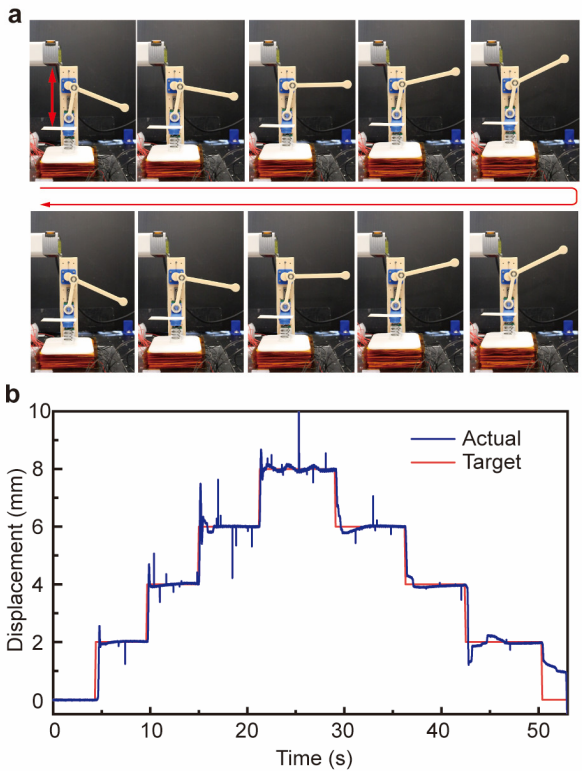

**Fig. 12.** The rotary joint developed based on the SE-HASEL actuator. (a) The snapshots of the rotary motion of the joint. (b) The displacement of SE-HASEL actuator.

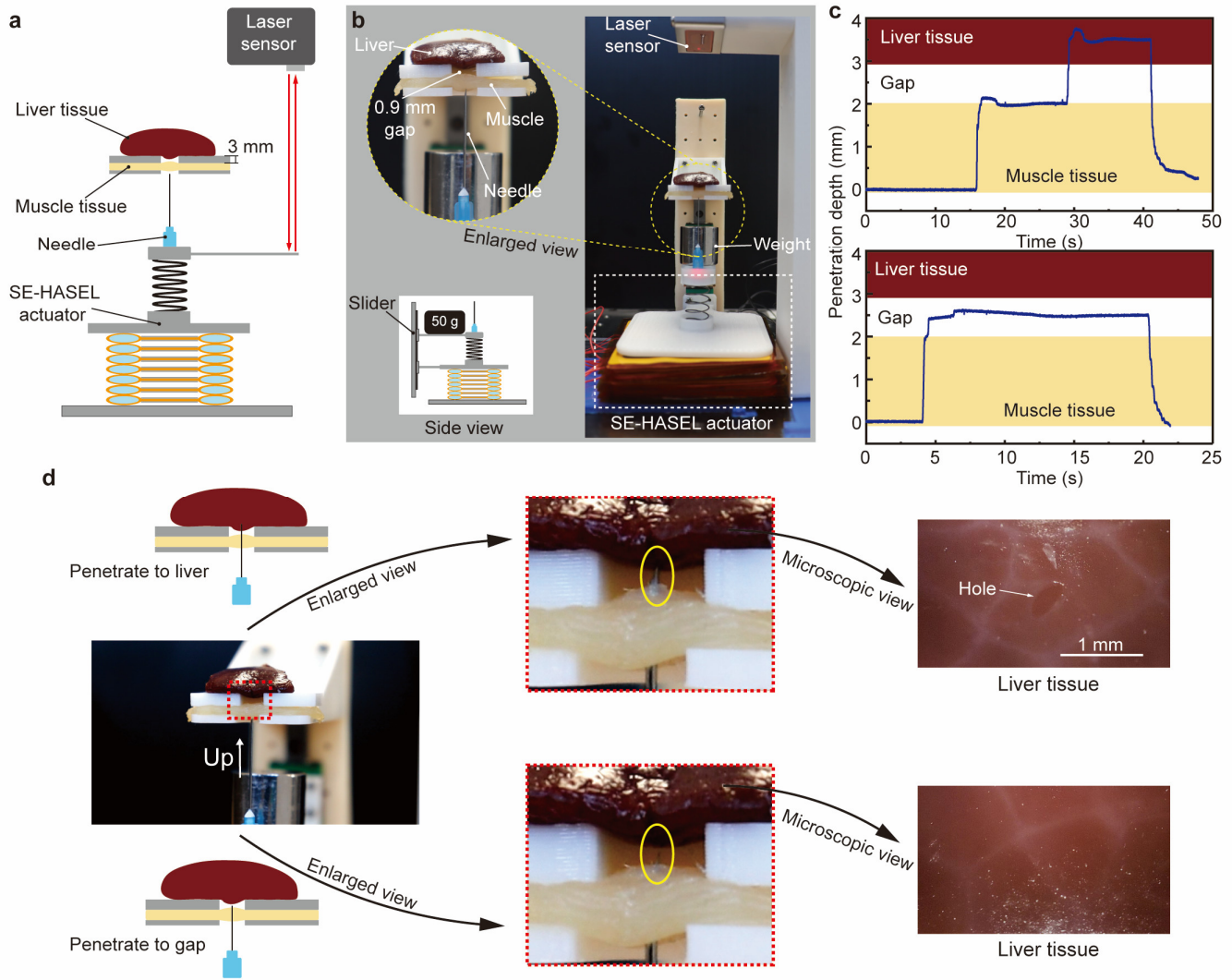

**Fig. 13.** The needle biopsy robot by the SE-HASEL actuator. (a) The design of the needle biopsy robot. (b) The actual assembled needle biopsy robot. (c) The actual displacement of SE-HASEL actuator. (d) The needle biopsy for sampling the liver tissue and the gap tissue. For sampling liver tissue, the needle penetrates to the liver with a pierced hole. For sampling gap tissue, the needle only penetrates to the gap.

V. CONCLUSION

In this article, our primary novelty revolved around using low-frequency AC square wave voltage to drive electrohydraulic actuators, effectively mitigating the gradual EA force decay observed under constant DC voltage. To further enhance their stability and precision, we introduced SE-HASEL actuators, which seamlessly integrated an elastic component to achieve smooth, oscillation-free displacement output. The implementation of feedback control overcomes hysteresis and ensures accurate, stable, and seamless displacement outputs under various conditions. This control strategy, combined with our novel actuator designs, offered a powerful toolset for harnessing the full potential of electrohydraulic actuators through stable and smooth displacement output. In addition to our practical findings, we presented a mathematical model that elucidated the underlying principles governing the behavior of electrohydraulic actuators under different voltage inputs. Our contributions extended beyond theoretical frameworks, as we introduced a cost-effective fabrication method for electrohydraulic actuators using polyimide (PI) materials, renowned for their mechanical robustness and thermal tolerance. This approach further streamlined the production process, making these actuators more accessible for broader applications. The experimental validation of our research findings provided compelling evidence of the feasibility and effectiveness of our proposed solutions. Through a series of experiments, we demonstrated the behavior of electrohydraulic actuators under varying voltage inputs, loads, and control strategies. These experiments underscored the practical applicability of our technology and control methods. Lastly, our study showcased the versatility and potential of SE-HASEL actuators by introducing a rotary joint that effectively converted linear displacement into rotational movement and a precise needle biopsy robot actuated by our SE-HASEL actuator. These two robotic applications created a novel way for MRI-compatible medical robotics.

Currently, the mathematical model's capacitance and resistance parameters are unknown, resulting in only qualitative analysis of the behavior of electrohydraulic actuators. In subsequent works, we will determine the capacitance and the resistance parameters quantitatively by fitting the theoretical input current derived through (2) and (6) and the actual input current acquired from the high-voltage supply device under a constant DC input voltage. Additionally, according to the mathematical model of displacement mentioned in Section II.C, we will comprehensively optimize the design parameters of the SE-HASEL actuator, such as the volume of dielectric liquid and the size of electrodes. For the needle biopsy robot, we will substitute all the ferromagenetic components (sliders, needle, spring) with components made of magnetic-compatible materials, and test with a real MRI mechanie in the future.

ACKNOWLEDGMENT

The authors acknowledge the assistance of the Department of Mechanical Engineering, National University of Singapore.

APPENDIX

The legends of supporting videos:
Video S1: The displacement output of the electrohydraulic actuators under DC voltage and AC voltage.
Video S2: The motion of Plate A and Plate B of the SE-HASEL actuator.
Video S3: Impact absorption of the SE-HASEL actuator
Video S4: The rotary robotic joint by the SE-HASEL actuator
Video S5: The needle biopsy robot by the SE-HASEL actuator

REFERENCES

[1] N. R. Sinatra, C. B. Teeple, D. M. Vogt, K. K. Parker, D. F. Gruber, and R. J. Wood, "Ultragentle manipulation of delicate structures using a soft robotic gripper," *Sci. Robot.*, vol. 4, no. 33, p. eaax5425, 2019.
[2] D. Rus and M. T. Tolley, "Design, fabrication and control of soft robots," *Nature*, vol. 521, no. 7553, pp. 467–475, 2015.
[3] S. Kim, C. Laschi, and B. Trimmer, "Soft robotics: A bioinspired evolution in robotics," *Trends Biotechnol.*, vol. 31, no. 5, pp. 287–294, 2013, doi: 10.1016/j.tibtech.2013.03.002.
[4] R. Coulson, C. J. Stabile, K. T. Turner, and C. Majidi, "Versatile soft robot gripper enabled by stiffness and adhesion tuning via thermoplastic composite," *Soft Robot.*, vol. 9, no. 2, pp. 189–200, 2022.
[5] H. K. Yap, H. Y. Ng, and C.-H. Yeow, "High-force soft printable pneumatics for soft robotic applications," *Soft Robot.*, vol. 3, no. 3, pp. 144–158, 2016.
[6] L. Rosalia et al., "Soft robotic patient-specific hydrodynamic model of aortic stenosis and ventricular remodeling," *Sci. Robot.*, vol. 8, no. 75, p. eade2184, 2023.
[7] D. Xie et al., "Fluid-Driven High-Performance Bionic Artificial Muscle with Adjustable Muscle Architecture," *Adv. Intell. Syst.*, p. 2200370, 2023.
[8] C. Tang et al., "A pipeline inspection robot for navigating tubular environments in the sub-centimeter scale," *Sci. Robot.*, vol. 7, no. 66, p. eabm8597, 2022.
[9] C. S. X. Ng, M. W. M. Tan, C. Xu, Z. Yang, P. S. Lee, and G. Z. Lum, "Locomotion of miniature soft robots," *Adv. Mater.*, vol. 33, no. 19, p. 2003558, 2021.
[10] S. Seok, C. D. Onal, K.-J. Cho, R. J. Wood, D. Rus, and S. Kim, "Meshworm: a peristaltic soft robot with antagonistic nickel titanium coil actuators," *IEEE/ASME Trans. mechatronics*, vol. 18, no. 5, pp. 1485–1497, 2012.
[11] C. Tawk, G. M. Spinks, M. in het Panhuis, and G. Alici, "3D printable linear soft vacuum actuators: their modeling, performance quantification and application in soft robotic systems," *IEEE/ASME Trans. Mechatronics*, vol. 24, no. 5, pp. 2118–2129, 2019.
[12] Y. Chi, Y. Li, Y. Zhao, Y. Hong, Y. Tang, and J. Yin, "Bistable and multistable actuators for soft robots: Structures, materials, and functionalities," *Adv. Mater.*, vol. 34, no. 19, p. 2110384, 2022.
[13] B. A. W. Keong and R. Y. C. Hua, "A Novel Fold-Based Design Approach toward Printable Soft Robotics Using Flexible 3D Printing Materials," *Adv. Mater. Technol.*, vol. 3, no. 2, pp. 1–13, 2018, doi: 10.1002/admt.201700172.
[14] R. Zhu et al., "Soft Robots for Cluttered Environments Based on Origami Anisotropic Stiffness Structure (OASS) Inspired by Desert Iguana," *Adv. Intell. Syst.*, p. 2200301, 2023.
[15] H. Zhao, K. O'brien, S. Li, and R. F. Shepherd, "Optoelectronically innervated soft prosthetic hand via stretchable optical waveguides," *Sci. Robot.*, vol. 1, no. 1, p. eaai7529, 2016.
[16] J. Zhu et al., "Intelligent soft surgical robots for next-generation minimally invasive surgery," *Adv. Intell. Syst.*, vol. 3, no. 5, p. 2100011, 2021.
[17] M. Cianchetti and A. Menciassi, "Soft robots in surgery," in *Soft Robotics: Trends, Applications and Challenges: Proceedings of the Soft Robotics Week, April 25-30, 2016, Livorno, Italy*, Springer, 2017, pp. 75–85.
[18] S. K. Mitchell et al., "An Easy-to-Implement Toolkit to Create Versatile and High-Performance HASEL Actuators for Untethered Soft Robots," *Adv. Sci.*, vol. 6, no. 14, 2019, doi: 10.1002/advs.201900178.
[19] L. T. Gaeta et al., "Magnetically induced stiffening for soft robotics," *Soft Matter*, vol. 19, no. 14, pp. 2623–2636, 2023.
[20] P. Rajagopalan, M. Muthu, Y. Liu, J. Luo, X. Wang, and C. Wan, "Advancement of Electroadhesion Technology for Intelligent and Self-Reliant Robotic Applications," *Adv. Intell. Syst.*, vol. 4, no. 7, p. 2200064, 2022.
[21] J. Guo, J. Leng, and J. Rossiter, "Electroadhesion technologies for robotics: A comprehensive review," *IEEE Trans. Robot.*, vol. 36, no. 2, pp. 313–327, 2019.
[22] D. Wei et al., "Electrostatic Adhesion Clutch with Superhigh Force Density Achieved by MXene-Poly(Vinylidene Fluoride-Trifluoroethylene-Chlorotrifluoroethylene) Composites," *Soft Robot.*, vol. 10, no. 3, pp. 482–492, 2023, doi: 10.1089/soro.2022.0013.
[23] Q. Xiong, B. W. K. Ang, T. Jin, J. W. Ambrose, and R. C. H. Yeow, "Earthworm-Inspired Multi-Material, Adaptive Strain-Limiting, Hybrid Actuators for Soft Robots," *Adv. Intell. Syst.*, vol. 5, no. 3, p. 2200346, 2023.
[24] J. Shintake, V. Cacucciolo, D. Floreano, and H. Shea, "Soft robotic grippers," *Adv. Mater.*, vol. 30, no. 29, p. 1707035, 2018.
[25] E. W. Schaler, D. Ruffatto, P. Glick, V. White, and A. Parness, "An electrostatic gripper for flexible objects," in *2017 IEEE/RSJ International Conference on Intelligent Robots and Systems (IROS)*, IEEE, 2017, pp. 1172–1179.
[26] V. Alizadehyazdi, M. Bonthron, and M. Spenko, "An electrostatic/gecko-inspired adhesives soft robotic gripper," *IEEE Robot. Autom. Lett.*, vol. 5, no. 3, pp. 4679–4686, 2020.
[27] Q. Xiong et al., "So-EAGlove: VR Haptic Glove Rendering Softness Sensation With Force-Tunable Electrostatic Adhesive Brakes," *IEEE Trans. Robot.*, vol. 38, no. 5, pp. 3450–3462, 2022.

[28] K. Zhang, E. J. Gonzalez, J. Guo, and S. Follmer, "Design and analysis of high-resolution electrostatic adhesive brakes towards static refreshable 2.5 D tactile shape display," *IEEE Trans. Haptics*, vol. 12, no. 4, pp. 470–482, 2019.

[29] R. J. Hinchet and H. Shea, "Glove-and Sleeve-Format Variable-Friction Electrostatic Clutches for Kinesthetic Haptics," *Adv. Intell. Syst.*, vol. 4, no. 12, p. 2200174, 2022.

[30] Y. Chen, S. Xu, Z. Ren, and P. Chirarattananon, "Collision resilient insect-scale soft-actuated aerial robots with high agility," *IEEE Trans. Robot.*, vol. 37, no. 5, pp. 1752–1764, 2021.

[31] Z. Ren et al., "A high-lift micro-aerial-robot powered by low-voltage and long-endurance dielectric elastomer actuators," *Adv. Mater.*, vol. 34, no. 7, p. 2106757, 2022.

[32] H. Zhao, A. M. Hussain, M. Duduta, D. M. Vogt, R. J. Wood, and D. R. Clarke, "Compact dielectric elastomer linear actuators," *Adv. Funct. Mater.*, vol. 28, no. 42, p. 1804328, 2018.

[33] J. H. Youn et al., "Dielectric elastomer actuator for soft robotics applications and challenges," *Appl. Sci.*, vol. 10, no. 2, 2020, doi: 10.3390/app10020640.

[34] E. Acome et al., "Hydraulically amplified self-healing electrostatic actuators with muscle-like performance," *Science (80-. )*., vol. 359, no. 6371, pp. 61–65, 2018, doi: 10.1126/science.aao6139.

[35] N. Kellaris, V. G. Venkata, G. M. Smith, S. K. Mitchell, and C. Keplinger, "Peano-HASEL actuators: Muscle-mimetic, electrohydraulic transducers that linearly contract on activation," *Sci. Robot.*, vol. 3, no. 14, 2018, doi: 10.1126/scirobotics.aar3276.

[36] E. L. Foster et al., "MR-compatible treadmill for exercise stress cardiac magnetic resonance imaging," *Magn. Reson. Med.*, vol. 67, no. 3, pp. 880–889, 2012.

[37] R. Gassert, A. Yamamoto, D. Chapuis, L. Dovat, H. Bleuler, and E. Burdet, "Actuation methods for applications in MR environments," *Concepts Magn. Reson. Part B Magn. Reson. Eng. An Educ. J.*, vol. 29, no. 4, pp. 191–209, 2006.

[38] K. Chinzei, R. Kikinis, and F. A. Jolesz, "MR compatibility of mechatronic devices: design criteria," in *Medical Image Computing and Computer-Assisted Intervention–MICCAI'99: Second International Conference, Cambridge, UK, September 19-22, 1999. Proceedings 2*, Springer, 1999, pp. 1020–1030.

[39] M. Rajendra et al., "Motion generation in MR environment using electrostatic film motor for motion-triggered cine-MRI," *IEEE/ASME Trans. Mechatronics*, vol. 13, no. 3, pp. 278–285, 2008.

[40] N. Kellaris et al., "Spider-Inspired Electrohydraulic Actuators for Fast, Soft-Actuated Joints," *Adv. Sci.*, vol. 8, no. 14, pp. 1–17, 2021, doi: 10.1002/advs.202100916.

[41] Z. Yoder, D. Macari, G. Kleinwaks, I. Schmidt, E. Acome, and C. Keplinger, "A Soft, Fast and Versatile Electrohydraulic Gripper with Capacitive Object Size Detection," *Adv. Funct. Mater.*, vol. 33, no. 3, 2023, doi: 10.1002/adfm.202209080.

[42] T. Nakamura and A. Yamamoto, "Modeling and control of electroadhesion force in DC voltage," *Robomech J.*, vol. 4, no. 1, pp. 1–10, 2017.

[43] F. Giraud, M. Amberg, and B. Lemaire-Semail, "Merging two tactile stimulation principles: electrovibration and squeeze film effect," in *2013 World Haptics Conference (WHC)*, IEEE, 2013, pp. 199–203.

[44] E. Leroy, R. Hinchet, and H. Shea, "Multimode hydraulically amplified electrostatic actuators for wearable haptics," *Adv. Mater.*, vol. 32, no. 36, p. 2002564, 2020.

[45] Y. Sun, K. M. Digumarti, H. V. Phan, O. Aloui, and D. Floreano, "Electro-Adhesive Tubular Clutch for Variable-Stiffness Robots," *IEEE Int. Conf. Intell. Robot. Syst.*, vol. 2022-Octob, pp. 9628–9634, 2022, doi: 10.1109/IROS47612.2022.9982098.

[46] Z. Yoder et al., "Design of a High-Speed Prosthetic Finger Driven by Peano-HASEL Actuators," *Front. Robot. AI*, vol. 7, no. November, pp. 1–17, 2020, doi: 10.3389/frobt.2020.586216.

[47] S. Kim and Y. Cha, "Electrohydraulic actuator based on multiple pouch modules for bending and twisting," *Sensors Actuators A Phys.*, vol. 337, no. February, p. 113450, 2022, doi: 10.1016/j.sna.2022.113450.

[48] S.-D. Gravert et al., "Low Voltage Electrohydraulic Actuators for Untethered Robotics," *arXiv Prepr. arXiv2306.00549*, 2023.

[49] D. J. Meyer, M. A. Peshkin, and J. E. Colgate, "Fingertip friction modulation due to electrostatic attraction," in *2013 world haptics conference (WHC)*, IEEE, 2013, pp. 43–48.

[50] Q. Xiong, X. Zhou, J. W. Ambrose, and R. C.-H. Yeow, "An Amphibious Fully-Soft Miniature Crawling Robot Powered by Electrohydraulic Fluid Kinetic Energy," Sep. 2023, doi: arXiv:2309.11020.

[51] C. F. Dalziel, "Electric shock hazard," *IEEE Spectr.*, vol. 9, no. 2, pp. 41–50, 1972.

[52] S. Doyle, M. Feldman, J. Tomaszewski, and A. Madabhushi, "A boosted Bayesian multiresolution classifier for prostate cancer detection from digitized needle biopsies," *IEEE Trans. Biomed. Eng.*, vol. 59, no. 5, pp. 1205–1218, 2010.

[53] K. Masamune et al., "Development of an MRI-compatible needle insertion manipulator for stereotactic neurosurgery," *J. Image Guid. Surg.*, vol. 1, no. 4, pp. 242–248, 1995.

[54] G. R. Sutherland, P. B. McBeth, and D. F. Louw, "NeuroArm: an MR compatible robot for microsurgery," in *International congress series*, Elsevier, 2003, pp. 504–508.

[55] M. Ho, Y. Kim, S. S. Cheng, R. Gullapalli, and J. P. Desai, "Design, development, and evaluation of an MRI-guided SMA spring-actuated neurosurgical robot," *Int. J. Rob. Res.*, vol. 34, no. 8, pp. 1147–1163, 2015.